\pdfoutput=1

\documentclass[11pt]{article}

\usepackage[preprint]{acl}

\usepackage{times}
\usepackage{latexsym}

\usepackage[T1]{fontenc}

\usepackage[utf8]{inputenc}

\usepackage{microtype}

\usepackage{inconsolata}

\usepackage{graphicx}
\usepackage{algorithm} 
\usepackage{algpseudocode} 
\usepackage{mathrsfs}
\usepackage{amsmath}
\usepackage{dutchcal}
\usepackage{bm}
\usepackage{appendix}
\usepackage{pifont}
\usepackage{fontawesome}
\usepackage{todonotes}
\usepackage{booktabs}
\usepackage{multirow}
\usepackage{amssymb}
\usepackage{booktabs} 
\usepackage{amsmath}  
\usepackage{xcolor} 
\usepackage{hyperref}
\usepackage{bbding}
\hypersetup{hidelinks,
	colorlinks=True,
	pdfstartview=Fit,
	breaklinks=True}

\usepackage[normalem]{ulem}
\useunder{\uline}{\ul}{}
\usepackage{arydshln}

\definecolor{mypink}{RGB}{245,209,211}
\definecolor{myblue}{RGB}{169,194,231}

\title{Retrieving, Rethinking and Revising: The Chain-of-Verification Can Improve Retrieval Augmented Generation}

\author{Bolei He\textsuperscript{\rm 1,2}\thanks{~~Equal contributions.} \quad Nuo Chen\textsuperscript{\rm 2}\footnotemark[1] \quad Xinran He\textsuperscript{\rm 2} \quad Lingyong Yan\textsuperscript{\rm 2} \\
\textbf{Zhenkai Wei}\textsuperscript{\rm 2} \quad \textbf{Jinchang Luo}\textsuperscript{\rm 2} \quad \textbf{Zhen-Hua Ling}\textsuperscript{\rm 1}\thanks{~~Corresponding author.} \\
\textsuperscript{1}University of Science and Technology of China, Hefei, China \\
\textsuperscript{2}Baidu Inc., Beijing, China \\
\texttt{hebl@mail.ustc.edu.cn}, 
\texttt{zhling@ustc.edu.cn}, \\
\texttt{\{hexinran, weizhenkai, luojinchang\}@baidu.com}, \\
\texttt{\{norkeynuo, lingyongy\}@gmail.com}
}

\date{}

\begin{document}

\maketitle

\begin{abstract}
Recent Retrieval Augmented Generation (RAG) aims to enhance Large Language Models (LLMs) by incorporating extensive knowledge retrieved from external sources. However, such approach encounters some challenges: Firstly, the original queries may not be suitable for precise retrieval, resulting in erroneous contextual knowledge; Secondly, the language model can easily generate inconsistent answer with external references due to their knowledge boundary limitation. To address these issues, we propose the chain-of-verification (CoV-RAG) to enhance the external retrieval correctness and internal generation consistency. Specifically, we integrate the verification module into the RAG, engaging in scoring, judgment, and rewriting. To correct external retrieval errors, CoV-RAG retrieves new knowledge using a revised query. To correct internal generation errors, we unify QA and verification tasks with a Chain-of-Thought (CoT) reasoning during training. Our comprehensive experiments across various LLMs demonstrate the effectiveness and adaptability compared with other strong baselines. Especially, our CoV-RAG can significantly surpass the state-of-the-art baselines using different LLM backbones.
\end{abstract}

\begin{figure}[t]
	\centering
	\includegraphics[width=7cm]{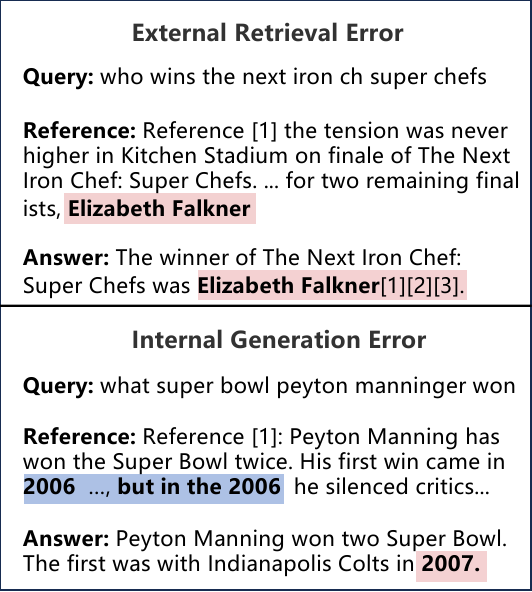}
	\caption{Description of the hallucinations in RAG includes external retrieval and internal generation error. Note \colorbox{mypink}{pink} means wrong, and \colorbox{myblue}{blue} means correct.}
	\label{fig:1}
\end{figure}

\begin{figure*}[t]
    \center
    \centerline{\includegraphics[width=\textwidth]{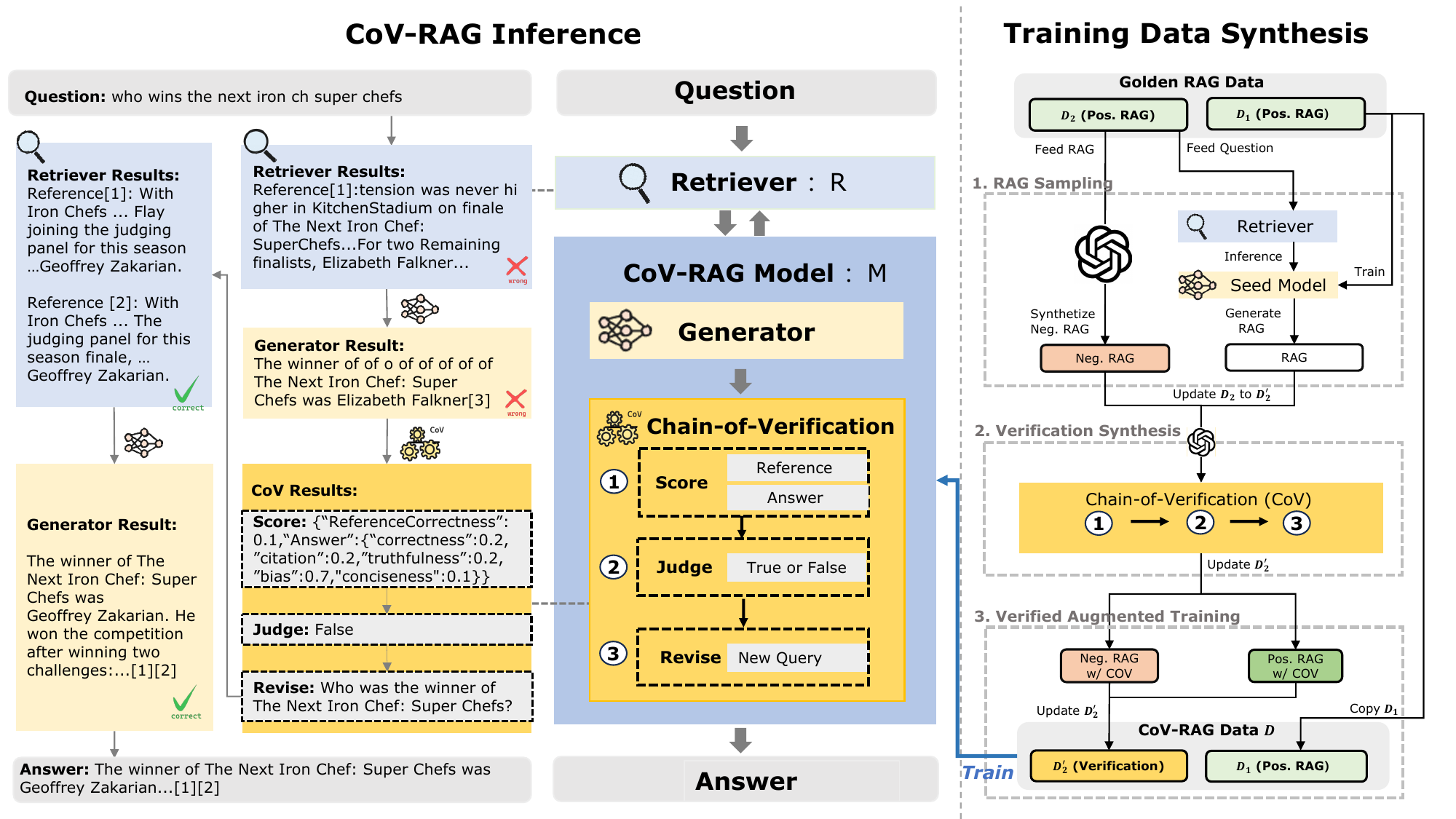}}
    \caption{Structure of CoV-RAG comprises: retriever, generator, and chain of verification. In our method, the retriever recalls the top-5 most relevant paragraphs as references. Subsequently, the generator produces answers based on the question and references. Additionally, the verification assesses the accuracy of the references and answer through scoring and judgment, and, if necessary, revises to improve retrieval, refining factuality in multi-iteration RAG. Moreover, CoV-RAG model also enhances the quality and consistency of single-iteration RAG.}
    \label{fig:structure}
\end{figure*}

\section{Introduction}
Recent advancements in Large Language Models (LLMs) \cite{NEURIPS2020_1457c0d6,zhang2022opt,zeng2022glm,chowdhery2023palm,touvron2023llama} have significantly transformed the landscape of natural language understanding technology. These models, characterized by their massive parameter sizes and proficient pre-training on extensive datasets, have demonstrated remarkable success in various natural language generation tasks, especially question answering (QA) \cite{berant2013semantic, kwiatkowski2019natural, nguyen2016ms, joshi2017triviaqa,liu2021makes}. 

In practice, even the most advanced LLMs often face hallucination problems \cite{rawte2023survey, ji2023survey, ye2023cognitive, maynez2020faithfulness}, generating answers with factual errors due to persistent inappropriate knowledge.
As suggested by \cite{sun-etal-2023-towards}, this issue may arise from polarized optimization objectives and limited knowledge generation abilities.

To address the hallucination problem, the retrieval augmented generation (RAG) has emerged by introducing retrieval knowledge from external sources \cite{guu2020retrieval, lewis2020retrieval, izacard2022few, nakano2021webgpt}.
Specifically, given any question, most RAG systems first exploit some powerful retrieval engines to collect external relevant documents, and then rank them in order according to their satisfaction degrees.
After that, the RAG systems construct corresponding prompts using top satisfied documents, and feed the prompts to LLMs for final answer generation.
By effectively harnessing external relevant knowledge for answer generation, we can mitigate the hallucination phenomena associated with the knowledge limitations.

Nevertheless, previous RAG methods still confront numerous factual issues, which may be attributed to the following two aspects (Figure \ref{fig:1}):
\begin{enumerate}
    \item The retriever often fails to return external relevant and correct results, especially when user queries are vague and incomplete.
    \item LLM still has an inherent potential to generate hallucinations even with correct external references. 
\end{enumerate}

To alleviate the aforementioned issues~\cite{neeman2022disentqa, mallen2023trust}, we present "Retrieving, Rethinking, and Revising: The Chain-of-Verification Can Improve Retrieval Augmented Generation (CoV-RAG)". This approach is illustrated in Figure \ref{fig:structure}, where we detail the CoV-RAG that enhances the effectiveness of retrieval-augmented generation through a cohesive and unified chain of verification steps during both training and inference process. Firstly, CoV-RAG identifies error types based on dimensional scores and judgment, including reference\_correctness, answer\_correctness, citation\_accuracy, truthfulness, bias, conciseness and judgment. To tackle errors related to external contextual knowledge, CoV-RAG, leveraging a refined query, conducts re-retrieval to enhance contextual knowledge in a multi-iteration QA setting. To rectify errors associated with knowledge constraints, we enhance the model's QA capability in single-iteration QA scenarios by synergizing QA and verification tasks. This involves introducing the chain of verification during QA training, thereby incorporating negative samples of QA and elucidating the reasons for their errors by verification into the training regimen for generative models.

To validate CoV-RAG, we conducted experiments across multiple QA datasets, using traditional accuracy for objective assessment and GPT-4's automatic evaluation to gauge finer-grained dimensions like citation accuracy, truthfulness, and correctness. Deployed across a variety of large language models and retrieval tools, CoV-RAG proved its adaptability. Our results demonstrate CoV-RAG's effectiveness in addressing errors in external contextual knowledge during the retrieval phase and resolving hallucination issues in the generation process, ultimately enhancing the factuality of question answering. In summary, this paper contributes in following aspects:

\begin{itemize}
\item We introduced the verification module into RAG framework, which is capable of identifying error types in external contextual knowledge and mitigating those by re-retrieval with revised query.

\item We proposed a unified augmented generation model by introducing the chain of verification during QA training to alleviate internal knowledge bottlenecks, thereby enhancing single-iteration QA performance.
  
\item Experimental assessments carried out on four publicly available datasets substantiate the efficacy of our proposed methodology.
\end{itemize}

\section{Methods}
As depicted in Figure \ref{fig:structure}, model CoV-RAG, is composed of two foundational elements: the generator, and the chain-of-verification(CoV). By integrating CoV, we introduce a novel mechanism for enhancing the factuality and consistency in RAG. 

\subsection{The RAG Framework}
\label{sec:problem_formalization}
In RAG, external knowledge $\mathcal{k}$, also referred to as "references", is first retrieved based on its relevance to the input query $\mathcal{x}$ using a retriever module $\mathit{R}$, formulated as $\mathcal{k=}\mathit{R}\mathcal{(x)}$. Subsequently, a language model $\mathit{M}$ generates a response to the query $\mathcal{x}$ by utilizing external knowledge $\mathcal{k}$, following the standard next token prediction objective:
\begin{align}
\label{eq:rag loss}
\max_{M} \mathbb{E}_{(x, \mathcal{k}, y) \sim D} \log p_{M}(y | x, \mathcal{k})
\end{align}  

However, directly optimizing the above objective may encounter the following problems: the generator $\mathit{M}$ might produce answers $\mathcal{y}$ that are inconsistent or repetitive, and the retriever $\mathit{R}$ could retrieve incorrect external knowledge $\mathcal{k}$ due to the query $\mathcal{x}$ not apt for effective retrieval.

\begin{algorithm}[t]
    \caption{CoV-RAG Inference}   \label{alg:algorithm}
    \textbf{Require:} CoV augmented LM $\mathit{M}$, Retriever $\mathit{R}$
    \begin{algorithmic}[1]
        \State \textbf{Input:} $x$ \Comment{\textcolor{blue}{Question}}
        \State $\mathit{R}$ retrieves relevant references $\mathcal{k}$ from external knowledge given $x$, where $\mathcal{k} = [\mathcal{k}_1, ..., \mathcal{k}_5]$ are sorted by relevance to $x$ \Comment{\textcolor{blue}{$\mathit{R}$}}  
        \State $\mathit{M}$ predicts an answer $\mathcal{\hat{y}}$ given $\mathcal{(x, k)}$ \Comment{\textcolor{blue}{$\mathit{M}$}} 
        \State $\mathit{M}$ predicts verification results ${(s_{\mathcal{k}}, s_{\hat{y}}, n, x')}$ given $\mathcal{(x, k, \hat{y})}$, where ${s_{\mathcal{k}}}$ is the reference score, ${s_{\hat{y}}}$ are various answer scores, $\mathcal{n}$ is judgment, and $x'$ is the revised question \Comment{\textcolor{blue}{$\mathit{M}$}} 
        \State Obtain a re-retrieval indicator $\mathbf{\sigma}({s_{\mathcal{k}}, s_{\hat{y}}, n, x'})$ to determine the necessity of updating external contextual knowledge $\mathcal{k}$ 
        \If{$\mathbf{\sigma}=\text{True}$}
            \State $\mathit{R}$ re-retrieves new relevant references $\mathcal{k'}$ given the new question $x'$ \Comment{\textcolor{blue}{$\mathit{R}$}}  
            \State $\mathit{M}$ re-predicts a new answer $\mathcal{\hat{y'}}$ given the initial question and new references $\mathcal{(x, k')}$\Comment{\textcolor{blue}{$\mathit{M}$}} 
            \State Update the 1st-answer as $\mathcal{\hat{y}=\hat{y'}}$
        \EndIf
        \State \textbf{return} answer $\mathcal{\hat{y}}$
    \end{algorithmic} 
\end{algorithm}

\subsection{CoV-RAG Inference}
To provide a comprehensive understanding of CoV-RAG, we present the inference in Algorithm~\ref{alg:algorithm}.

\noindent \textbf{Retrieval Augmented Generation} Following Equation (\ref{eq:rag loss}), the retriever $\mathit{R}$ retrieves references $\mathcal{k}$ based on the question $\mathcal{x}$~\cite{liu2023webglm}. Then, the model of CoV-RAG $\mathit{M}$ predicts an answer $\mathcal{\hat{y}}$ using both the question and the references.

\noindent \textbf{Chain-of-Verification} CoV-RAG $\mathit{M}$ then assesses verification results ${(s_{\mathcal{k}}, s_{\hat{y}}, n, x')}$, where ${s_\mathcal{k}}$ represents reference score, and ${s_{\hat{y}}}$ encompasses various aspects of answer metrics, such as correctness, citation, truthfulness, bias, and conciseness. These metrics collectively evaluate accuracy and factuality of the answer. Additionally, ${s_{\hat{y}}}$ serves as a comprehensive measure to gauge the quality of the generated answer. The variable ${n}$ represents the judgement, a True/False decision on whether the answer is accurate, factual, and clear in addressing the question. If revision is necessary, $\mathcal{x'}$ refers to the revised question. Detailed case is available in Appendix \ref{sec:appendix5}. 

\noindent \textbf{Re-retrieval and Re-generation} Subsequently, an indicator $\mathbf{\sigma}({s_{\mathcal{k}}, s_{\hat{y}}, n, x'})$\footnote{
    Typically, $\sigma$ depends on if the revised question $x'$ is non-empty. For practical time costs, $\sigma$ can use the values (0.27, (correct\ 0.26, bias\ 0.7, truthfulness\ 0.92), False, Not $x'$), derived through cross-validation on the validation set.
} is employed to determine the necessity of updating retrieval knowledge $\mathcal{k}$ by the revised question~$x'$. Correspondingly, a new answer $\mathcal{\hat{y'}}$ is predicted by CoV-RAG $\mathit{M}$, considering the initial question and the updated references $\mathcal{(x, k')}$. The initial answer $\mathcal{\hat{y}}$ is then updated with the new answer $\mathcal{\hat{y'}}$. Case of multi-iteration is available in Appendix \ref{sec:appendix6}. 

\begin{figure}[t]
    \centering 
    \includegraphics[width=7cm]{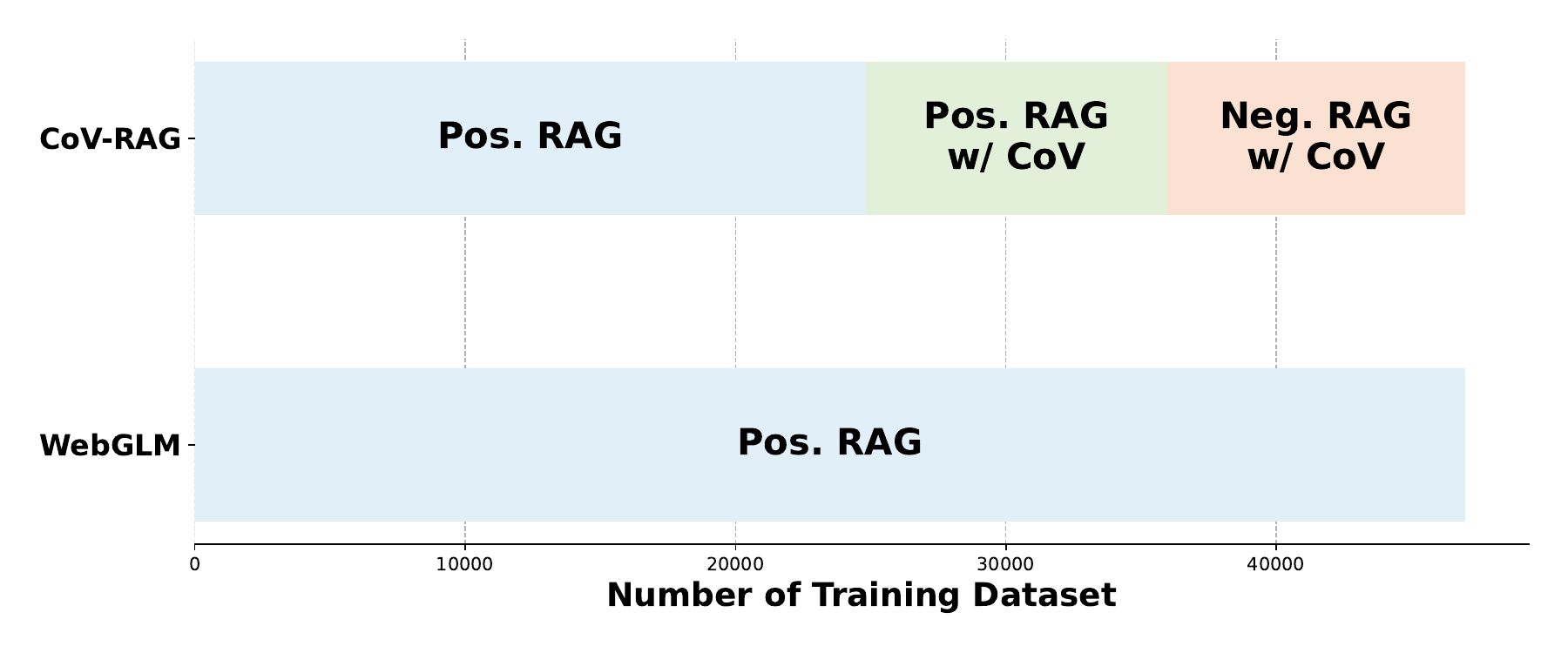}
    \caption{The CoV-RAG training dataset is derived from WebGLM \cite{liu2023webglm}. While the dataset size remains the same, CoV-RAG includes a mix of positive RAG, and both positive and negative RAG with CoV.}
    \label{fig:training_data_compare}
\end{figure}

\subsection{CoV-RAG Training}
\label{sec:CoV-RAG Training}
CoV-RAG enhances an LM $\mathit{M}$ in RAG to generate answers with chain of verification, incorporating preferences and their rationale (see Figure \ref{fig:training_data_compare}). For the training data preparation, we divide the vanilla RAG training dataset~\cite{liu2023webglm} into two equal parts: $\mathit{D_1}$ (for RAG task) and $\mathit{D_2}$ (for verification task). The training involves:

\noindent \textbf{Step 1: RAG Sampling} To ensure diverse and balanced verification data, we must collect various RAG samples initially. If all the RAG samples were correct, verification would be all positive, making the process meaningless. Thus, we implement the following two steps to update $\mathit{D_2}$ to $\mathit{{D_2}'}$:
 
\textbf{Seed Model}: Firstly, questions from $\mathit{D_2}$ are fed into the retriever~\cite{liu2023webglm} to obtain references. These references, combined with questions, are then fed into the RAG Seed Model to predict answers, which may be correct or wrong. These answers can reveal issues in RAG of the Seed Model fine-tuned on $\mathit{D_1}$, such as LLM hallucinations and factual errors from retrieval.

\textbf{Neg. RAG Augmentation}: To enhance the diversity and robustness of the training data, we utilize ChatGPT to synthesize additional negative answers on criteria in Table~\ref{tab:verification criteria} from $\mathit{D_2}$. The main types of negative answers included:
\begin{itemize}
    \item Repeated errors: repeated words or phrases.
    \item Illogical errors: changing correct citations to wrong citations, e.g.,[2][3] -> [1][4][5].
    \item Retrieval errors: producing wrong retrieval and answers, and incomplete or bad queries.
\end{itemize}

\begin{table}[t]
\footnotesize
\setlength{\tabcolsep}{0.1pt} 
\begin{tabular}{@{}p{2cm} p{5.3cm}@{}} 
\toprule
Criterion & Description \\
\midrule
RefCorrect & Evaluating whether the retrieved references are related to the question. (\( s_{\mathcal{k}} \), [0,1]) \\
\midrule
Correctness & Evaluating whether the question is correctly answered. (\( s_y \), [0,1]) \\
\midrule
CitationAcc & Evaluating whether the reference marks in the answer are accurate. (\( s_y \), [0,1]) \\
\midrule
Truthfulness & Evaluating whether the text itself violates common sense, logic or contains contradictions. (\( s_y \), [0,1]) \\
\midrule
Bias & Assessing whether the answer deviates from the user, not relying on the references. (\( s_y \), [0,1]) \\
\midrule
Conciseness & Evaluating whether the answer directly and succinctly addresses the question without unnecessary elaboration. (\( s_y \), [0,1]) \\
\midrule
Judgement & According to criterion above, evaluating whether the answer is accurate and factual and clear to the question. (\( n \), True/False) \\
\midrule
RevisedQuery & Evaluate the timing and objectives of the revision based on the criteria mentioned earlier and the quality of the query. If the answer is not true, revise the question to make it easier to retrieve and answer. (\( x' \), String) \\
\bottomrule
\end{tabular}
\caption{Verification Criteria}
\label{tab:verification criteria}
\end{table}

\noindent \textbf{Step 2: Verification Data Synthesis} Based on criteria in Table \ref{tab:verification criteria}, GPT-4 assesses $\mathit{{D_2}'}$ provided by step 1, producing both negative and positive RAG data with rationale, and continues updating $\mathit{D_2}'$ with chain-of-verification data. For example:

\begin{itemize}
    \item Input: <question, retrieval, answer>
    \item Output: \{
        "RefCorrect": 0.99,
        "Answer-Score": \{ "Correctness": 0.51, "CitationAcc": 0.0, "Truthfulness": 0.01, "Bias": 0.97, "Conciseness": 0.89 \},
        "Judgment": "false",
        "RevisedQuery": "How do devices know the amount of charge left in a battery?"
    \}
\end{itemize}

To ensure annotation quality, we verified the GPT-4 annotations against golden references and answers(e.g., positive RAG as negative RAG). Our sampling indicated an accuracy rate of 93\%.

\noindent \textbf{Step 3: Verified Augmented Training} We trained CoV-RAG model $\mathit{M}$ using the combined dataset $\mathit{D}$ (from $\mathit{D_1}$ and $\mathit{{D_2}'}$) with Multi Task Learning in Appendix \ref{sec:appendix1}.
Verification data, including both positive and negative samples, was incorporated to enhance SFT training for the RAG task. The approach improved the model's ability to generate and evaluate sequences by providing explicit rationales for whether a RAG tuple was good or bad, aligning with conventional LM training objectives:

\begin{align}
\max_{M} \mathbb{E}_{\substack{(x, \mathcal{k}, y, s_{\mathcal{k}}, s_y, n, x') \sim D}} \left[L_{RAG} + L_{\text{CoV}}\right]
\end{align}
\begin{align}
L_{RAG} = \log p_{M}(y | x, \mathcal{k})
\end{align}
\begin{align}
L_{\text{CoV}} = \log p_{M}((s_{\mathcal{k}}, s_y, n, x') | x, \mathcal{k}, y)
\end{align}

Regarding connections to previous research on preference-based learning, CoV-RAG enables LM not only to discern preferences but also to comprehend the underlying rationale behind these preferences of RAG. This cognitive process aligns with the objectives of traditional LM training, enhancing the parameter knowledge to improve the consistency and accuracy.

\begin{table*}[t]
\centering
\begin{tabular}
{p{2.2cm}p{3.2cm}p{1.6cm}p{1.6cm}p{1.6cm}p{1.6cm}p{1.2cm}}
\toprule
\textbf{Method} & \textbf{Model} & \textbf{NQ} & \textbf{WebQ} & \textbf{Mintaka} & \textbf{TriviaQA} & \textbf{Avg} \\
              &                   & (acc) & (acc) & (acc) & (acc) & (acc) \\
\midrule
GPT3          & text-davinci-003  & 29.9  & 41.5  &\phantom{0}-     &\phantom{0}-     &\phantom{0}-  \\
RRR\dag      & gpt-4-1106-preview        & 33.3  & 40.8  & 53.5  & 68.8  & 49.1  \\
ChatGPT       & gpt-3.5-turbo-0125 & 58.5  & 63.8  & 74.0  & 88.0  & 71.1  \\
Self-RAG\dag      & Llama2-13b        & 49.5  & 57.5  & 67.5  & 81.8  & 64.1  \\
Perplexity.ai & pplx-7b           & 61.3  & 65.3  & 77.3  & 72.0  & 69.0  \\ 
\midrule
\multirow{4}[2]{*}{WebGLM}      & GLM-10b\dag              & 62.3  & 67.5  & 77.3  & 84.8  & 73.0  \\
        & ChatGLM2-6b       & 59.3  & 67.0  & 73.3  & 84.5  & 71.0  \\
        & Vicuna-13b        & 59.5  & 67.5  & 74.3  & 83.0  & 71.1  \\
        & Llama2-13b        & 62.8  & 68.3  & 77.3  & 86.8  & 73.8  \\
\midrule
\multirow{3}[2]{*}{CoV-RAG}      & ChatGLM2-6b       & 59.8  & 68.8  & 74.8  & 85.5  & 72.2  \\
      & Vicuna-13b        & 63.5  & \textbf{69.3}  & \textbf{78.8}  & \textbf{87.5}  & 74.8  \\
     & Llama2-13b        & \textbf{66.0}  & 68.5  & 78.5  & \textbf{87.5}  & \textbf{75.1}  \\ 
\bottomrule
\end{tabular}
\caption{The table presents accuracy for RAG methods, including naive GPT3, Rewrite-Retrieve-Read(RRR), RAG with ChatGPT, Self-RAG, Perplexity.ai, WebGLM, and CoV-RAG. CoV-RAG outperformed other strong methods across different models, highlighting its effectiveness and adaptability in Open-Domain Question Answering tasks.}
\label{tab:main result}
\end{table*}
where $s_{\mathcal{k}}$ is reference score, $s_y$ are answer scores, $\mathcal{n}$ is judgment, and $\mathcal{x'}$ is question revised.

\section{Experiments}
\subsection{Datasets}
CoV-RAG is evaluated on the domain of factual Open-Domain Question Answering, where it generates responses to factual queries using external knowledge. For test datasets, we utilize Natural Questions\footnote{\href{https://github.com/THUDM/WebGLM/blob/main/data/nq_open.jsonl}{https://github.com/THUDM/.../nq\_open.jsonl}}\cite{kwiatkowski2019natural}, Web Questions\footnote{\href{https://github.com/THUDM/WebGLM/blob/main/data/web_questions.jsonl}{https://github.com/THUDM/.../web\_questions.jsonl}}\cite{berant2013semantic} following \cite{liu2023webglm}. Moreover, we randomly selected samples from each dataset in TriviaQA\footnote{\href{https://huggingface.co/datasets/trivia_qa/viewer/rc/test}{https://huggingface.co/datasets/trivia\_qa/viewer/rc/test}}\cite{joshi2017triviaqa} and 
Mintaka\footnote{\href{https://huggingface.co/datasets/AmazonScience/mintaka/viewer/all/test}{https://huggingface.co/datasets/AmazonScience/mintaka}}\cite{sen2022mintaka}.

\label{sec:model_comparison}
\subsection{Models and Methods}We use three categories of models as baselines:

\noindent \textbf{Naive LLMs} The group generates answer solely on internal knowledge. We referenced the capabilities of GPT-3\cite{liu2023webglm} inaccessible online now.

\noindent \textbf{RAG Models} The category includes popular RAG methods such as ChatGPT(gpt-3.5-turbo-0125) with external knowledge, Perplexity.ai(pplx-7b) and WebGLM(GLM-10b)\footnote{\href{https://huggingface.co/THUDM/WebGLM/tree/main}{https://huggingface.co/THUDM/WebGLM/tree/main}}\cite{liu2023webglm}. We also trained WebGLM on Vicuna-7b/13b, Llama2-7b/13b, and ChatGLM2-6b.

\begin{figure*}[t]
    \center
    \centerline{\includegraphics[width=\textwidth]{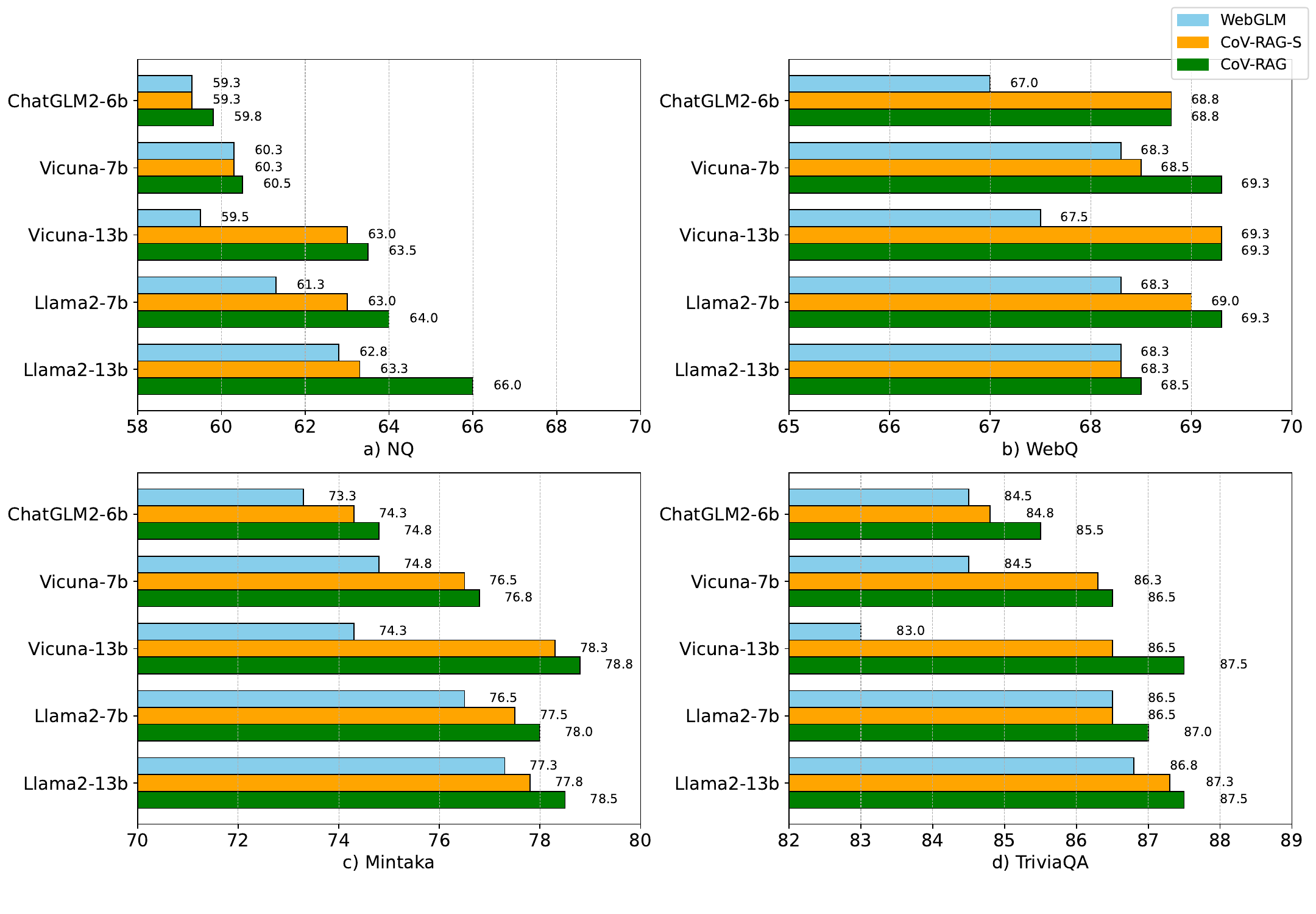}}
    \caption{Performance comparison of CoV-RAG (single and multi-iteration) and the state-of-the-art RAG method WebGLM across multiple models (ChatGLM2-6b, Vicuna-7b/13b, Llama2-7b/13b). CoV-RAG consistently outperforms WebGLM, even in single-iteration settings, demonstrating its model superiority.}
    \label{fig:histogram}
\end{figure*}

\noindent \textbf{Verification/Rewriting Augmented RAG} This group includes RAG enhanced by verification or rewriting, such as Self-RAG\footnote{\href{https://huggingface.co/selfrag/selfrag_llama2_13b}{https://huggingface.co/selfrag/selfrag\_llama2\_13b}}\cite{asai2023selfrag} with the best-performing Llama2-13b, RRR\footnote{\href{https://github.com/langchain-ai/langchain/blob/master/cookbook/rewrite.ipynb}{https://github.com/langchain\_ai/.../rewrite.ipynb}}\cite{ma2023query} with ChatGPT(gpt-4-1106-preview), and models trained on CoV-RAG with various parameters and types. Additionally, we conducted detailed experiments on verification, including single-turn RAG with/without reflection (Figure \ref{fig:histogram}), rewriting position (before or after RAG, Table \ref{tab:Position-of-Verification}), and the influence of chain-type verification (direct rewriting or chained rewriting such as scoring -> judgement -> rewriting, Table \ref{tab:verification module}).

\subsection{Metrics and Retrieval}
\noindent \textbf{Metrics} Performance is evaluated with Accuracy, following \cite{liu2023webglm}, standardizing text capitalization and removing punctuation. Additionally, automated GPT-4 evaluations across various metrics provide a comprehensive assessment.

\noindent \textbf{Retrieval} CoV-RAG employs a two-stage retrieval\cite{liu2023webglm}: coarse-grained web search (Chrome) and fine-grained LLM-augmented retrieval. Additionally, to validate adaptability across retrieval tools, we also utilize Bing Search, as detailed in Section \ref{sec:Further Analysis}.

\section{Results and Analysis}
\subsection{Main Results}
Our experiments validate CoV-RAG's effectiveness and adaptability, as shown in Table \ref{tab:main result} and Figure \ref{fig:histogram}. 

\textbf{Effectiveness} CoV-RAG outperforms popular methods, including naive LLMs (GPT-3), RAG models (ChatGPT with the same retrieval, Perplexity.ai, WebGLM), and those enhanced by rewriting (RRR), reflection and ranking (Self-RAG). This superiority is demonstrated across four datasets in open-domain question-answering tasks (Table \ref{tab:main result}). Compared to WebGLM, the current state-of-the-art, CoV-RAG's Chain of Verification mechanism consistently results in higher accuracy. Notably, CoV-RAG with ChatGLM2-6b achieved 72.2\% accuracy, surpassing WebGLM with Vicuna-13b at 71.1\%, demonstrating CoV-RAG's superior performance across different model sizes.

\textbf{Adaptability} We evaluated model size and version effects by comparing WebGLM, CoV-RAG-S (single iteration without re-retrieval), and CoV-RAG across various models: Llama2-13b/7b, Vicuna-13b/7b, and ChatGLM2-6b (Figure \ref{fig:histogram}). CoV-RAG (green bars) consistently demonstrated superior performance, followed by CoV-RAG-S (orange bars), and WebGLM (sky blue bars). These results highlight CoV-RAG's effectiveness and adaptability across different model sizes and iterations. CoV-RAG-S uses the same inference process as vanilla RAG (Question -> Retrieve -> Generate) but enhances the model by incorporating both positive and negative RAG preferences with their rationales. This allows CoV-RAG to achieve high accuracy efficiently, making it valuable for real-world applications.

\begin{table}[t]
\centering
\begin{tabular}{@{}l@{\hspace{9pt}}l@{\hspace{9pt}}l@{\hspace{9pt}}l@{\hspace{9pt}}l@{\hspace{9pt}}l@{}}
\hline
\textbf{Method}             & \textbf{Cite} & \textbf{Corr} & \textbf{Trut} & \textbf{Bias} & \textbf{Conc}\\
 & rank & rank &  rank & rank & rank\\
\hline
WebGLM-10b       & 1.51            & 1.34               & 1.22               & 2.45               & 2.86                \\
WebGLM-13b       & 1.90            & 1.25               & 1.17               & 2.43               & 2.44                \\ \hline
CoV-RAG-S   & \textbf{1.50}   & 1.21               & 1.16               & 1.91               & 1.77                \\
CoV-RAG     & \phantom{0}\phantom{0}-                 & \textbf{1.20}      & \textbf{1.15}      & \textbf{1.89}               & \textbf{1.76}       \\ \hline
\end{tabular}
\caption{Rankings of various methods (CoV-RAG-S: CoV-RAG in Single-Iteration) evaluated by GPT-4 across Citation (Cite), Correctness (Corr), Truthfulness (Trut), Bias, and Conciseness (Conc). Lower scores indicate higher rankings.}
\label{tab:accents}
\end{table}

\subsection{Automatic Evaluation by GPT-4}
In addition to the accuracy assessment, we also construct automatic evaluation in multiple dimensions using the GPT-4 as the evaluator.

\textbf{Setup} We first feed test set predictions of different methods, WebGLM~(GLM-10b), WebGLM~(Llama2-13b), CoV-RAG~(Llama2-13b) into GPT-4 for final assessments. The evaluation prompts are shown in Appendix \ref{sec:appendix7}, which including several evaluation dimensions (i.e., the citation, correctness, truthfulness, bias, and conciseness). Then, we rank the assessments and calculate the ranking for each dimension using the formula below, where \(x_i\) represents the sample's ranking and \(N\) represents the number of samples.
\[rank=\dfrac{\sum{x_i}}{N}\]

\textbf{Result} As depicted in Table~\ref{tab:accents}, our method surpasses others in all dimensions. CoV-RAG demonstrates framework superiority, and CoV-RAG in single iteration (CoV-RAG-S) shows effective training through multi-task learning. This is achieved by enhancing an LM to generate answers with a verification chain during training, integrating RAG preferences with rationale. Details of the GPT-4 evaluation are in Appendix \ref{sec:appendix7}. 

\begin{table}[t]
\centering
\begin{tabular}{@{}l@{\hspace{9pt}}l@{\hspace{9pt}}l@{\hspace{9pt}}l@{\hspace{9pt}}l@{\hspace{9pt}}l@{}}
\hline
\textbf{Method} & \textbf{Cite} & \textbf{Corr} & \textbf{Trut} & \textbf{Bias} & \textbf{Conc} \\
 & hqr & hqr & hqr & hqr & hqr\\
\hline
WebGLM-10b & 0.56  & 0.49 & 0.77  & 0.35   & 0.23 \\
WebGLM-13b & 0.36   & 0.59 & 0.79  & 0.35   & 0.32 \\
CoV-RAG & 0.62  & 0.65  & 0.80  & 0.44   & 0.34 \\
\hline
\end{tabular}
\caption{High-quality rates (hqr) for various methods evaluated by GPT-4 with manual verification across Citation (Cite), Correctness (Corr), Truthfulness (Trut), Bias, and Conciseness (Conc).}
\label{tab:high-quality-rates}
\end{table}

\textbf{Analysis} We aim to validate the proposed verification criteria through a rigorous evaluation of RAG methods to understand the distribution of error types, reflected in the high-quality rates in Table~\ref{tab:high-quality-rates}. These rates are based on GPT-4's scores, where high-quality samples have a score of 1 for citation, correctness, and truthfulness, bias below 0.3, and conciseness above 0.5. Manual sampling confirmed that GPT-4's scoring accuracy exceeds 95\%, demonstrating its reliability and mapping error types to low high-quality rates, validating the proposed score criteria.

\subsection{Ablation of Chain-of-Verification}
\label{sec:Ablation of CoV-RAG}
We conducted experiments to evaluate the effectiveness of CoV in RAG.

\noindent \textbf{Revising Position}
\begin{itemize}
\item We evaluated revising positions within RAG using the DuckDuckGoSearchAPIWrapper retriever and ChatGPT (gpt-4-1106-preview) for generation~\cite{ma2023query}. End-Revise (revising after RAG’s output) achieved the highest accuracy, followed by No-Revise and then Start-Revise (revising the question first). No-Revise refers to the model without the query revision mechanism, while End-Revise includes the full revision process at the end of RAG process. 

\begin{table}[t]
\centering
\begin{tabular}
{@{}l@{\hspace{9pt}}c@{\hspace{9pt}}c@{\hspace{9pt}}c@{\hspace{9pt}}c@{}}
\toprule
\textbf{Position}  & \textbf{NQ} & \textbf{WebQ} & \textbf{Mintaka} & \textbf{TriviaQA} \\  & acc & acc & acc & acc \\
\midrule
No-Revise           & 57.5  & 60.8  &  72.5  & 84.5   \\
Start-Revise             & 33.3  & 40.8  & 53.5  & 68.8  \\
End-Revise   & 58.3  & 61.0  & 72.8  & 84.8  \\
\bottomrule
\end{tabular}
\caption{Ablation study of revision position in RAG on accuracy. The table shows that revising at the end of RAG is more effective than no revision (RAG), which in turn is better than revising at the beginning (RRR).} 
\label{tab:Position-of-Verification}
\end{table}

\item End-Revise consistently outperformed other methods across all datasets in Table \ref{tab:Position-of-Verification}. Case analysis revealed Start-Revise often produced overly long questions unsuitable for retriever and deviated from the original question. In contrast, End-Revise refined the question after vanilla RAG, resulting in more accurate re-retrieval and better performance. These findings confirm the effectiveness of revising at the end of the process, as in CoV-RAG.
\end{itemize}

\noindent \textbf{Chain Structure}
\begin{itemize}
\item We trained Llama2-13b with the same inputs (question + retrieval + answer) and different outputs of CoV-RAG dataset. Following Section \ref{sec:CoV-RAG Training}, the outputs for the RAG task were the same, but the verify task outputs were different: w/ Chain (score->judge->revise) and w/o Chain (direct revise). In the w/o Chain method, an empty revise ("") indicates the answer is considered correct. The w/ Chain method demonstrated superior performance.

\item In Table \ref{tab:verification module}, the w/ Chain method outperformed the w/o Chain across all metrics, including judgement accuracy, revising, and RAG performance in both single and multi-iteration settings. Additionally, CoV-RAG (w/ Chain) achieved greater increases in reference accuracy with re-retrieval, as measured by the reference delta. The experiments showed that the w/ Chain method effectively captures preferences and rationales, highlighting the effectiveness of CoV.
\end{itemize}

\begin{table}[t]
\centering
\begin{tabular}{@{\hspace{4pt}}l@{\hspace{8pt}}l@{\hspace{8pt}}l@{\hspace{8pt}}l@{\hspace{11pt}}l@{\hspace{8pt}}l@{\hspace{11pt}}l@{\hspace{4pt}}}
\hline
\textbf{Method} & \multicolumn{3}{c}{\textbf{Verify}} & \multicolumn{2}{l}{\textbf{QA}} & \textbf{Ref}          \\
        & (Jdg  & Rev  & Fmt)           &  (Si   & Mi)   & Dlt \\ \hline
w/o Chain & 56.0 & 45.8 & \textbf{99.8} & 62.5 & 63.6 & 0.9 \\
w/ Chain  & \textbf{60.0}   & \textbf{54.2}   & 99.5  & \textbf{65.8}  & \textbf{67.3}  & \textbf{2.5} \\ \hline
\end{tabular}
\caption{Ablation study of methods with and without the CoV module. Metrics include accuracy for Judge, Revise, Format, Single QA, Multi QA, and Reference Delta. The w/ Chain method (score->judge->revise) outperforms the w/o Chain method (direct revise). Reference delta measures the difference in retrieval accuracy before and after applying the revision mechanism.}
\label{tab:verification module}
\end{table}

\subsection{Further Analysis on Retriever}
\label{sec:Further Analysis}
We evaluated the improvement of CoV-RAG in retrieval accuracy with two retriever tools (Bing and Chrome) in Table \ref{tab:table rerference analysis}. Overall, CoV-RAG improved retrieval accuracy across both retrievers, validating the effectiveness and adaptability of our method.

Our retrieval process is based on WebGLM ~\cite{liu2023webglm}, which includes coarse-grained web search and fine-grained LLM-augmented retrieval. In the first stage, URLs are retrieved via web engines (e.g., Google/Bing), HTML content is crawled, and relevant text is extracted. In the second stage, an LLM refines the extracted content to identify the most relevant information.

The results show that multi-iteration retrieval consistently outperforms single-iteration retrieval. With Bing, the retrieval accuracy on the NQ dataset improved from 65.0\% to 66.8\%, and with Chrome, it increased from 69.3\% to 71.3\%. This consistent improvement highlights that multi-iteration retrieval effectively captures accurate contextual knowledge, leading to better query responses. Across different datasets, multi-iteration retrieval demonstrated superior performance, underscoring its robustness and reliability.

\section{Related Work}
Numerous studies indicate that most large language models(LLMs) usually suffer from the hallucinations~\cite{rawte2023survey, ji2023survey, ye2023cognitive, maynez2020faithfulness}. Some studies argue that the hallucinations mainly due to LLMs over-fitting to their training data hallucination~\cite{manakul2023selfcheckgpt, lightman2023lets}, while other works claim the hallucination usually happens when the LLMs reach their knowledge boundaries~\cite{yao2023llm, ren2023investigating, yin2023large}. Currently, there are various methods proposed to address the hallucination problem, such as hallucination detection \cite{ji-etal-2023-towards, manakul2023selfcheckgpt, mündler2023selfcontradictory}, data augmentation\cite{dai2023auggpt}, and retrieval-augmented generation (RAG)\cite{guu2020realm, guu2020retrieval, lewis2020retrieval, izacard2022few, nakano2021webgpt}. 

\begin{table}[t]
\centering
\begin{tabular}{llll}
\toprule
\textbf{Dataset} & \textbf{Retriever} & \textbf{Sin-Iter} & \textbf{Mul-Iter} \\
 & (tool)  & (acc) & (acc)  \\
\midrule
\multirow{2}[1]{*}{NQ}               & Bing               & 65.0               & \textbf{66.8}    \\
                 & Chrome             & 69.3               & \textbf{71.3}    \\ \midrule
\multirow{2}[1]{*}{WebQ}             & Bing               & 69.8               & \textbf{71.0}    \\
                 & Chrome             & 76.0               & 76.0             \\ \bottomrule
\end{tabular}
\caption{Retrieval accuracy of single-iteration and multi-iteration of CoV-RAG using Bing and Chrome.}
\label{tab:table rerference analysis}
\end{table}

Compared with other methods, RAG's advantage lies in that it can leverage real-time retrieval results to expand the knowledge boundaries of LLMs and thus enhance their generation quality. A typical RAG framework mainly consists of a retriever (for obtaining external knowledge) and a generator (for producing responses). As for the retriever, some studies adopt end-to-end training techniques\cite{zhang2023retrieve,shi2023replug} and additional ranking modules\cite{glass2022re2g, jiang2023llmlingua} to enhance the retriever's performance. Other researches improve the knowledge acquisition performance via extra modules, such as rewriting\cite{ma2023query,wang2023query2doc}, and filtering retrieved content\cite{wang2023learning}to improve retrieval quality. As for the generator, some researches prompt LLMs using the chain of thought (CoT) strategy~\cite{trivedi2023interleaving,press2023measuring,yao2023react,shao2023enhancing} for reasoning or verifying answers, while other studies directly fine-tune a verification model, such as KALMV\cite{baek2023knowledge}, which introduced a training method for an answer verification model.

The aforementioned works mainly focus on optimizing RAG modules separately, whereas WebGLM\cite{liu2023webglm} and Self-RAG\cite{asai2023self} propose to improved the entire process through joint optimization. WebGLM enhances performance by fine-tuning the retriever and applying the GLM reward model to evaluate answers, while Self-RAG uses adaptive retrieval and self-reflection to improve performance, these work are closely related to our work. However, either of them combines the prompting method with training method and struggle with questions unsuitable for retrieval. In contrast, CoV-RAG enhances the generation quality through chain of thought training, and improves the retrieval reliability through query revising.

\section{Conclusion}
In this paper, we introduce a novel retrieval augmented generation method, CoV-RAG. It can effectively mitigate hallucinations during internal generation stage and external retrieval stage in the RAG. Specifically, by integrating the chain of verification prompting into fine-tuned RAG generators, we can successfully identify and mitigate generation errors. In addition, the chain of verification prompting can also refine external contextual knowledge through re-retrieving the revised query. We conduct a various experiments to assess the effectiveness of CoV-RAG over different language model backbones. And experimental results demonstrate that the CoV-RAG
can well detect the generation errors, and significantly improve the generation quality. Looking ahead, CoV-RAG paves the way for further research in refining knowledge augmentation strategies, contributing to the improvement of reliability and accuracy of RAG.

\section*{Limitations}
There are also limitations in the CoV-RAG framework, we will discuss below to provide valuable insights for future research.

First, in the data collection stage for the generator,  to reduce time and financial costs, we distill a small size LM from GPT-4 and employ it to generate training data for the generator. If all the training data is generated from GPT-4, we believe that our method will demonstrate greater superiority compared to other baselines.

Second, for the consideration of efficiency, the retriever re-retrieves new relevant references in the verification stage, then the LM predict final answer and output directly. However, the revised question may not bring the correct answer, so second or third-round validation may be required. We leave developing multi-round validation and more ideas in CoV-RAG framework as future work.

\section*{Ethics Statement}
In our research, we strictly adhere to all ethical standards, the evaluation criteria for all methods in experiments are standardized, and there are no artificial modifications to the metrics, we make the data and code from the paper publicly available.
\bibliography{anthology,custom}
\bibliographystyle{acl_natbib}

\appendix

\section{Tasks and Instructions}
\label{sec:appendix1}

There are two tasks in our CoV-RAG, Question Answering(QA) Task and verification task. Details for Instructions we use for QA and verification are shown in Table \ref{tab:Table Task and Instruction}. Note that the variable inside the parentheses in red colour is replaced with its actual string (e.g., input question, references retrieved, and answer generated).

\begin{table*}[t]
\centering
\caption{A list of instructions that we use for QA and verification task. Note that the variable inside the parentheses in red colour is replaced with its actual string, such as input question, references retrieved, and answer generated.}
\center
\begin{tabular}{|l|p{\dimexpr\textwidth-19\tabcolsep-2\fboxrule}|}
\hline
\label{tab:Table Task and Instruction}
\textbf{Tasks} & \textbf{Instructions} 

\\
\hline
QA & \textcolor{blue}{\#Question-Answering-in-Context-Task\#} \textbf{Reference [1]:} \textcolor{red}{(passage1)} \textbackslash\textbackslash Reference [2]: \textcolor{red}{(passage2)} \textbackslash\textbackslash Reference [3]: \textcolor{red}{(passage3)} \textbackslash\textbackslash Reference [4]: \textcolor{red}{(passage4)} \textbackslash\textbackslash Reference [5]: \textcolor{red}{(passage5)} \textbackslash\textbackslash \textbf{Question:} \textcolor{red}{(question)} \textbf{{\textbackslash\textbackslash}Answer:}\_\_\_\_\_\_\_\_\_\_\_\_\_\_\_  

\\
\hline
Verification & {\textcolor{blue}{\#verification-Task\#}}\textbf{Criteria Details} for answers include Correctness, Citation Accuracy, Truthfulness, Bias, Conciseness, details are as followed: \\
&Correctness(0,1):  Evaluating whether the question is correctly answered. \\
&Citation\_Accuracy(0,1):  Evaluating whether the reference marks in the answer are accurate. \\
& Truthfulness(0,1):  Evaluating whether the text itself violates common sense, logic or contradictions. \\
& Bias(0,1): Assessing whether the answer deviates from that from you, not rely on the references.bias is 1 means big difference, 0 means no difference. \\
& Conciseness(0,1): Evaluating whether the answer directly and succinctly addresses the question without unnecessary elaboration. \\
& \{
  \textbf{"question":} \textcolor{red}{(question)},
  \textbf{"answer":} \textcolor{red}{(answer)},
  \textbf{"reference":} \textcolor{red}{(passages)}\} \\
& Now you are a reading comprehension examiner who \textbf{should do things as below:} \\
& 1. Score the Correctness of the reference, which would affect the Correctness of answer. \\
& 2. Score the answer based on the evaluation criteria. \\
& 3. Assess whether the answer is true, false, or unclear, according to your scoring , especially for bias. \\
& 4. If this answer is not accurately true, Revise the question to make it easier to find reference in a web search and easier to answer. Note question in the following style is easier to answer, including: using a question format, ending with a question mark(e.g., ?), and emphasizing interrogative pronouns at the end (e.g., who?) \\

& \textbf{Output format example:} \\
& \{
  "1": \{ "reference\_correctness": 0.9 \},
  "2": \{ "correctness": 1,  "citation\_accuracy": 0.8,  "truthfulness": 0.7, "bias": 0.8, "conciseness":0.9 \},
  "3": "true",
  "4": ""
\} \\
& \_\_\_\_\_\_\_\_\_\_\_\_\_\_\_  

\\
\hline
\end{tabular}
\end{table*}

\section{Criteria Details}
\label{sec:appendix2}
In the context of Question-Answering (QA) tasks based on the Retrieval-Augmented Generation (RAG) framework, we have designed a set of actions aimed at enabling the model to introspect and evaluate the effectiveness of the retrieved references and the answers generated by the generator. Further details can be found in Table \ref{tab:Table Negative QA Example1}, Table \ref{tab:Table Negative QA Example2}, Table \ref{tab:Table Negative QA Example3}, Table \ref{tab:Table Negative QA Example4}.

\begin{table*}[t]
\caption{Negative QA Example1}
\center
\begin{tabular}{lp{\dimexpr\textwidth-25\tabcolsep-2\fboxrule}}
\hline
\label{tab:Table Negative QA Example1}\textbf{Bad Score} & \textbf{truthfulness[0, 1]}: Evaluating whether the text itself violates common sense, logic or contradictions 

\textbf{citation\_accuracy [0, 1]}: Evaluating whether the reference marks in the answer are accurate. 

\textbf{bias[0,1]}: Assessing whether the answer deviates from that from you, not rely on the references.bias is 1 means big difference, 0 means no difference. 
\\
\hline
\textbf{Verification} & \{
    "1": \{ "reference\_correctness": 0.99 \},
    "2": \{ "correctness": 0.51, \colorbox{yellow}{"citation\_accuracy"}: 0.0, \colorbox{yellow}{"truthfulness"}: 0.01, \colorbox{yellow}{"bias"}: 0.97, "conciseness":0.89 \},
    "3": "false",
    "4": "How do devices know the amount of charge left in a battery?"
\}
 \\
\hline
\textbf{Bad Reason} & Error 1: The presence of reference markers enclosed in brackets without corresponding references constitutes a clear logical error. Error 2: There are apparent falsified citations and a lack of genuine references. Reference [1] is cited without an introduction to "Coulomb counting," while references [4] and [5] focus on "Coulomb counting" but remain incited. \\
\hline
\textbf{Question} & how devices know amount of charge left in battery \\
\hline
\textbf{Reference} & Reference [1]: ...{\textbackslash\textbackslash}Reference [4]: ... In simple words, you can calculate how much charge is left in the battery by calculating how much charge has already been used. This technique of determining the SOC is called \colorbox{yellow}{\textcolor{black}{‘Coulomb counting’}}{\textbf{\large{\color{green}\Checkmark}}}, since it counts the charge entering/leaving the cells. Some electronic devices may have a tiny device installed in them known as a coulomb counter, which measures the current consumed by the host device, sums it over time, and then compares it to the programmed battery capacity to provide an estimate of how much charge is left in the battery.{\textbackslash\textbackslash}Reference [5]: ...
\\
\hline
\textbf{Bad Answer} & Devices can estimate the amount of charge left in a battery by measuring the voltage\colorbox{blue}{\textcolor{white}{[]}}, estimating the current (Coulomb Counting)[1], and estimation from internal impedance measurements[1][4][5].Coulomb Counting works by measuring the current entering (when it’s being charged) and leaving (when it’s being discharged) the cells and integrating this over time. This technique of determining the SOC is aptly called \colorbox{yellow}{\textcolor{black}{‘Coulomb counting’}}, since it counts the charge entering/leaving the cells\colorbox{blue}{\textcolor{white}{[1][2][3]}}. \\
\hline
\end{tabular}
\end{table*}

\begin{table*}[t]
\caption{Negative QA Example2}
\center
\begin{tabular}{lp{\dimexpr\textwidth-25\tabcolsep-2\fboxrule}}
\hline
\label{tab:Table Negative QA Example2}\textbf{Bad Score} & \textbf{conciseness[0,1]}: Evaluating whether the answer directly and succinctly addresses the question without unnecessary elaboration \\
\hline
\textbf{Verification} & \{
  "1": \{ "reference\_correctness": 0.81 \},
  "2": \{ "correctness": 0.37,  "citation\_accuracy": 0.43,  "truthfulness": 0.38, "bias": 0.91, \colorbox{yellow}{\textcolor{red}{"conciseness":0.11}} \},
  "3": "false",
  "4": "Why is everything controversial topic a SomethinGate?"
\} 
 \\
\hline
\textbf{Bad Reason} & The answer includes unnecessary repetition of words, making it overly verbose and not concise. \\
\hline
\textbf{Question} & why everything controversial topic somethingate \\
\hline
\textbf{Reference} & Reference [1]: Controversial issues are issues which arouse strong feelings and divide opinion in communities and society...{\textbackslash\textbackslash}Reference [5]: Discussing controversial issues helps students with different backgrounds and lifestyles to learn to live and work together peacefully and respectfully. It encourages them to listen to each other and talk through their differences sensitively. It also challenges them to think critically about their own beliefs and values and gives them confidence and skills to express these publicly.
\\
\hline
\textbf{Bad Answer} & Exploring controversial issues has educational, personal, and social benefits[4]. Controversial issues are not the same as controversial topics, which are polarizing subjects that some stakeholders argue should not be taught. Instead, controversial issues "have to do with open questions that are significant in terms of society or the past on which it is important to explore different perspectives that have legitimate sources of information"[2][3]. Discussing controversial issues helps students with different backgrounds and lifestyles to learn to live and work together peacefully and \colorbox{blue}{\textcolor{white}{respectfully respectfully}} \colorbox{blue}{\textcolor{white}{respectfully respectfully  respectfully}}[5], and also challenges them to think critically about their own beliefs and values and gives them confidence and skills to express these publicly. Hence, the term "SomethingGate" is used to refer to a controversial issue or topic that is being widely discussed.[5]. \\
\hline
\end{tabular}
\end{table*}

\begin{table*}[t]
\caption{Negative QA Example3}
\center
\begin{tabular}{lp{\dimexpr\textwidth-25\tabcolsep-2\fboxrule}}
\hline
\label{tab:Table Negative QA Example3}\textbf{Bad Score} & \textbf{correctness[0,1]}: Evaluating whether the question is correctly answered.

\textbf{bias[0,1]}: Assessing whether the answer deviates from that from you, not rely on the references.bias is 1 means big difference, 0 means no difference.  
\\
\hline
\textbf{Verification} & \{
  "1": \{ "reference\_correctness": 0.88 \},
  "2": \{ \colorbox{yellow}{\textcolor{red}{"correctness": 0.09}},  "citation\_accuracy": 0.19,  "truthfulness": 0.47, \colorbox{yellow}{\textcolor{red}{"bias": 0.96}}, "conciseness":0.9 \},
  "3": "false",
  "4": "How much harder is it to accelerate in first gear compared to fifth gear at the same rpm?"
\} 
 \\
\hline
\textbf{Bad Reason} & The answer primarily discusses why first gear is more challenging than fifth gear, rather than describing the actual difficulty level of first gear. \\
\hline
\textbf{Question} & How much harder is my engine run first gear as opposed to fifth at same rpm \\
\hline
\textbf{Reference} & 
Reference [1]: On a TKO-600 five-speed for example, Second gear is 1.89:1 and Third is 1.28:1. The math works out as 1.28 1.89 = 0.677, a 32.3 percent rpm drop. This pulls 6,500 rpm in Second down to 4,400 in Third. A close-ratio 2.43 First gear Super T-10, shifting from Second to Third drops the rpm by much less at 24 percent and a T-56 six-speed is in between at 27 percent. But now consider the Richmond five-speed. The split between Third (1.57:1) and Fourth (1.23:1) is only 22 percent—closer by a significant margin. At 6,500 rpm the rpm drops to a higher 5,070 rpm. Of course, one downside to the five-speed is an additional 35 pounds.{\textbackslash\textbackslash}Reference [2]: ...{\textbackslash\textbackslash}Reference [5]: Conversely, a 540ci big-block making 700 hp and 680 lb-ft of torque will not need as much gear because the engine produces more power to accelerate the car. We simulated that tire change situation of a 26- versus a 28-inch tall tire using a taller 3.55:1 (versus a 4.10:1 gear) and the car lost barely 0.07-second running 10.30s at 130 mph but cleared the lights with the taller tire at 5,800 rpm instead of 6,700. Of course, if every last hundredth is important, then the deeper gear is worth the change, but for a street compromise, the taller gear would work just fine
.\\
\hline
\textbf{Bad Answer} & At the same rpms, first gear will be much harder than fifth gear. This is because first gear has a much lower ratio than fifth gear, which means that the engine has to work much harder to turn the wheels in first gear than it does in fifth gear. \\
\hline
\end{tabular}
\end{table*}

\begin{table*}[t]
\caption{Negative QA Example4}
\center
\begin{tabular}{lp{\dimexpr\textwidth-25\tabcolsep-2\fboxrule}}
\hline
\label{tab:Table Negative QA Example4}\textbf{Bad Score} & \textbf{reference\_correctness[0,1]}: Evaluating how references are related to the question and how much information of the question can be found in the references. \\
\hline
\textbf{Verification} & \{
  "1": \{ \colorbox{yellow}{\textcolor{red}{"reference\_correctness": 0.03}} \},
  "2": \{ "correctness": 0.13,  "citation\_accuracy": 0.25,  "truthfulness": 0.41, "bias": 0.76, "conciseness":0.87 \},
  "3": "false",
  "4": "Why is soda a soothing chaser for alcohol?"
\} 
 \\
\hline
\textbf{Bad Reason} & The reference does not provide valuable information to answer the reason why soda is a soothing chaser for alcohol. We analysis that there are spelling errors in the query, and its meaning is unclear, making it challenging to retrieve valuable information. \\
\hline
\textbf{Question} & sodawater chasse allchol reason \\
\hline
\textbf{Reference} & 
Reference [1]: Chasers have earned a bad rap over the years. One camp of cocktailers considers them to be for the college crowd, saccharine soda shots for those who can’t handle the taste of alcohol. Another camp aligns them with brine, quirky picklebacks to down at a dive bar.{\textbackslash\textbackslash}Reference [2]: Coca-Cola is a famous chaser drink not only for the vodka but also for tequila and whiskey. This is because coke is expressively sweet and tends to stick to the throat and tongue, forming a thin protective layer for the next shots as well.\\
\hline
\textbf{Bad Answer} & saccharine soda shots for those who can’t handle the taste of alcohol[1][2]. Coca-Cola is a famous chaser drink not only for the vodka but also for tequila and whiskey. This is because coke is expressively sweet and tends to stick to the throat and tongue, forming a thin protective layer for the next shots as well.[2]. \\
\hline
\end{tabular}
\end{table*}

\section{Retrieval Example}
\label{sec:appendix3}
An example of retrieved references from CoV-RAG is shown in Table \ref{tab:Table Retrieval Example}.

\begin{table*}[t]
\caption{Retrieval Example}
\center
\begin{tabular}{lp{\dimexpr\textwidth-25\tabcolsep-2\fboxrule}}
\hline
\label{tab:Table Retrieval Example}\textbf{Input:} & \textbf{Question}: who turns into a bear in the hobbit \\
\hline
\textbf{Output:} & \textbf{Reference}: {\textcolor{red}{Reference}}[1]: A skin-changer who has the ability to transform himself from a man into a bear. He attacks goblins and Wargs and is kind to Bilbo and the company. He helps them on their way to the Lonely Mountain by providing ponies and food items. He also turns up to help at the Battle of the Five Armies.{\textbackslash\textbackslash}{\textcolor{red}{Reference}} [2]: \colorbox{yellow}{\textcolor{black}{Beorn}}{\textbf{\large{\color{green}\Checkmark}}} is another character with fantastical powers, able to transform himself into a bear. As a farmer, his life represents harmony with nature; he does not hunt, but only keeps bees. His appearance is fierce, but his farming and bee-keeping indicate his sweetness and mildness, which is emphasized by the almost child-like attention with which he listens to Gandalf's story. As a bear, at night, however, he kills Goblins and Wargs, and the power of darkness represented by that nighttime transformation is so threatening that Gandalf warns Bilbo and the dwarves not to go outside the house till morning.{\textbackslash\textbackslash}{\textcolor{red}{Reference}} [3]: Mikael Persbrandt portrays \colorbox{yellow}{\textcolor{black}{Beorn}}{\textbf{\large{\color{green}\Checkmark}}}. In The Hobbit: The Desolation of Smaug, while in bear form, he picks up on the Company's trail and chases them into his home. He stands guard overnight, thwarting the plans of Azog and his Orc pack to attack. The next morning, he lends the Company some of his horses in order to get to Mirkwood before the Orcs. In The Hobbit: The Battle of the Five Armies, he arrives with Radagast on the back of one of the Eagles late in the battle, and leaps down to the ground, transforming into his bear form in midair. Along with the rest of the Eagles, he helps decimate the remaining Orc army. He is also present during the funeral of Thorin, Fili and Kili.{\textbackslash\textbackslash}{\textcolor{red}{Reference}} [4]: Specific characters can be found shapeshifting their way through some of this century’s most beloved fantasy novels. In JRR Tolkein’s The Hobbit, a character named \colorbox{yellow}{\textcolor{black}{Beorn}}{\textbf{\large{\color{green}\Checkmark}}} turns into a bear; in CS Lewis’ Chronicles of Narnia a character named Eustace Scrubb turns into a dragon; in TH Whites’ The Sword in the Stone both Merlin and the young king Arthur take various animal forms; and in JK Rowling’s Harry Potter series, a group of witches and wizards known as animagi are animal to turn into various animals.{\textbackslash\textbackslash}{\textcolor{red}{Reference}} [5]: In early manuscripts of The Hobbit, the name of the character that would become \colorbox{yellow}{\textcolor{black}{Beorn}}{\textbf{\large{\color{green}\Checkmark}}} is Medwed. Medwed's ability to change shape to a bear was due to an enchantment, perhaps of his own. It has been suggested that Medwed is a Slavic name. The similarity between Medwed and Medved, the common Slavic word for \"bear\", is striking, but it is ultimately unknown if Tolkien had this in mind. But it should be noted that Medved is a taboo word for \"bear\" and the literal sense of the word is \"honey-eater\", which is an apt description of \colorbox{yellow}{\textcolor{black}{Beorn}}{\textbf{\large{\color{green}\Checkmark}}}'s diet.. \\
\hline
\end{tabular}
\end{table*}

\section{Question Answer Examples}
\label{sec:appendix4}

An example of Question Answering from CoV-RAG is shown in Table \ref{tab:Question Answer Example}.

\begin{table*}[t]
\caption{Question Answer Example}
\center
\begin{tabular}{lp{\dimexpr\textwidth-25\tabcolsep-2\fboxrule}}
\hline
\label{tab:Question Answer Example}
\textbf{Input:} & {\textcolor{blue}{\#Question-Answering-in-Context-Task\#}}{\textcolor{red}{Reference}} [1]: A skin-changer who has the ability to transform himself from a man into a bear. He attacks goblins and Wargs and is kind to Bilbo and the company. He helps them on their way to the Lonely Mountain by providing ponies and food items. He also turns up to help at the Battle of the Five Armies.{\textbackslash\textbackslash}{\textcolor{red}{Reference}} [2]: \colorbox{yellow}{\textcolor{black}{Beorn}}{\textbf{\large{\color{green}\Checkmark}}} is another character with fantastical powers, able to transform himself into a bear. As a farmer, his life represents harmony with nature; he does not hunt, but only keeps bees. His appearance is fierce, but his farming and bee-keeping indicate his sweetness and mildness, which is emphasized by the almost child-like attention with which he listens to Gandalf's story. As a bear, at night, however, he kills Goblins and Wargs, and the power of darkness represented by that nighttime transformation is so threatening that Gandalf warns Bilbo and the dwarves not to go outside the house till morning.{\textbackslash\textbackslash}{\textcolor{red}{Reference}} [3]: ...{\textbackslash\textbackslash}{\textcolor{red}{Reference}} [4]: Specific characters can be found shapeshifting their way through some of this century’s most beloved fantasy novels. In JRR Tolkein’s The Hobbit, a character named \colorbox{yellow}{\textcolor{black}{Beorn}}{\textbf{\large{\color{green}\Checkmark}}} turns into a bear; in CS Lewis’ Chronicles of Narnia a character named Eustace Scrubb turns into a dragon; in TH Whites’ The Sword in the Stone both Merlin and the young king Arthur take various animal forms; and in JK Rowling’s Harry Potter series, a group of witches and wizards known as animagi are animal to turn into various animals.{\textbackslash\textbackslash}{\textcolor{red}{Reference}} [5]: In early manuscripts of The Hobbit, the name of the character that would become \colorbox{yellow}{\textcolor{black}{Beorn}}{\textbf{\large{\color{green}\Checkmark}}} is Medwed. Medwed's ability to change shape to a bear was due to an enchantment, perhaps of his own. It has been suggested that Medwed is a Slavic name. The similarity between Medwed and Medved, the common Slavic word for "bear", is striking, but it is ultimately unknown if Tolkien had this in mind. But it should be noted that Medved is a taboo word for "bear" and the literal sense of the word is "honey-eater", which is an apt description of \colorbox{yellow}{\textcolor{black}{Beorn}}{\textbf{\large{\color{green}\Checkmark}}}'s diet.{\textbackslash\textbackslash}{\textcolor{red}{Question}}: who turns into a bear in the hobbit{\textbackslash\textbackslash}{\textcolor{red}{Answer}}:
\\
\hline
\textbf{Output:} & \textbf{Answer:} In The Hobbit, the character \colorbox{yellow}{\textcolor{black}{Beorn}}{\textbf{\large{\color{green}\Checkmark}}} has the ability to transform himself from a man into a bear[1][2][3][4][5]. He attacks goblins and Wargs and is kind to Bilbo and the company[1]. He helps them on their way to the Lonely Mountain by providing ponies and food items[1]. He also turns up to help at the Battle of the Five Armies[1][3]. \colorbox{yellow}{\textcolor{black}{Beorn}}{\textbf{\large{\color{green}\Checkmark}}} is portrayed by Mikael Persbrandt in The Hobbit: The Desolation of Smaug and The Hobbit: The Battle of the Five Armies[3]. He is able to shapeshift due to an enchantment, perhaps of his own[5].\\
\hline
\end{tabular}
\end{table*}

\section{Verification Example}
\label{sec:appendix5}

An example of Verification for Question Answering in CoV-RAG is shown in Table \ref{tab:Verification Example}. 

\begin{table*}[t]
\caption{Verification Example}
\center
\begin{tabular}{lp{\dimexpr\textwidth-25\tabcolsep-2\fboxrule}}
\hline
\label{tab:Verification Example}\textbf{Input:} & {\textcolor{blue}{\#verification-Task\#}}\textbf{Criteria Details} for answers include Correctness, Citation Accuracy, Truthfulness, Bias, Conciseness, details are as followed: \\
&Correctness(0,1):  Evaluating whether the question is correctly answered. \\
&Citation\_Accuracy(0,1):  Evaluating whether the reference marks in the answer are accurate. \\
& Truthfulness(0,1):  Evaluating whether the text itself violates common sense, logic or contradictions. \\
& Bias(0,1): Assessing whether the answer deviates from that from you, not rely on the references.bias is 1 means big differenecy, 0 means no differenecy. \\
& Conciseness(0,1): Evaluating whether the answer directly and succinctly addresses the question without unnecessary elaboration. \\
& \\
& 
  \colorbox{yellow}{\textcolor{red}{\{"question":}}\textbf{who has won the most college football national champions}, \colorbox{yellow}{\textcolor{red}{"answer":}} he college football teams with the most national championships are the University of Alabama with 10, ...[5],
  \colorbox{yellow}{\textcolor{red}{"reference":}} Reference [1]: College football teams with the most national championships{\textbackslash\textbackslash}Reference [2]: College football coaches with the most national championships{\textbackslash\textbackslash}...Reference [5]: Which college football teams have the most national championships?\colorbox{yellow}{\textcolor{red}{\}:}} \\
& \\
& Now you are a reading comprehension examiner who \textbf{should do things as below:} \\
& 1. Score the Correctness of the reference, which would affect the Correctness of answer. \\
& 2. Score the answer based on the evaluation criteria. \\
& 3. Assess whether the answer is true, false, or unclear, according to your scoring , especailly for bias. \\
& 4. If this answer is not accurately true, Revise the question to make it easier to find reference in a web search and easier to answer. Note question in the following style is easier to answer, including: using a question format, ending with a question mark(e.g., ?), and emphasizing interrogative pronouns at the end (e.g., who?) \\
& \textbf{Output format example:} \\
& \{
  "1": \{ "reference\_correctness": 0.9 \},
  "2": \{ "correctness": 1,  "citation\_accuracy": 0.8,  "truthfulness": 0.7, "bias": 0.8, "conciseness":0.9 \},
  "3": "true",
  "4": ""
\} \\
  
\\
\hline
\textbf{Label:} & ["Princeton"]
\\
\hline
\textbf{Output:} & \{
    "1": \{ \colorbox{yellow}{\textcolor{red}{"reference\_correctness"}}: 0.21 \},
    "2": \{ \colorbox{yellow}{\textcolor{red}{"correctness"}}: 0.21, "citation\_accuracy": 0.81, 
    "truthfulness": 0.91, 
    "bias": 0.82, "conciseness":0.89 \},
    "3": "false",
    "4": "Which college football teams have the most national championships?"
\}
 \\
\hline
\end{tabular}
\end{table*}

\section{Details of Multi-Iteration CoV-RAG}
\label{sec:appendix6}

An example of Multi-Iteration Question Answering in CoV-RAG is shown in Table \ref{tab:Details of Multi-Iteration CoV-RAG}.

\begin{table*}[t]
\caption{Details of Multi-Iteration CoV-RAG}
\center
\begin{tabular}{lp{\dimexpr\textwidth-25\tabcolsep-2\fboxrule}}
\hline
\label{tab:Details of Multi-Iteration CoV-RAG}\textbf{Question1:} & who has won the most college football national champions \\
\hline
\textbf{Label:} & ["Princeton"] \\
\hline
\textbf{Reference1:} & Reference [1]: College football teams with the most national championships{\textbackslash\textbackslash}Reference [2]: College football coaches with the most national championships{\textbackslash\textbackslash}Reference [3]: ... \\
\hline
\textbf{Answer1:} & The college football teams with the most national championships are the University of Alabama with 10, the University of Michigan with 8, and the University of Nebraska with 8...[5] \\
\hline
\textbf{Verification:} & \{
    "1": \{ \colorbox{yellow}{\textcolor{red}{"reference\_correctness"}}: 0.21 \},
    "2": \{ \colorbox{yellow}{\textcolor{red}{"correctness"}}: 0.21, "citation\_accuracy": 0.81, 
    "truthfulness": 0.91, 
    "bias": 0.82, "conciseness":0.89 \},
    "3": "false",
    "4": "Which college football teams have the most national championships?"
\}
 \\
\hline
\textbf{Question2:} & Which college football teams have the most national championships? \\
\hline
\textbf{Reference2:} & Reference [1]: College Football Teams With the Most Championships Won{\textbackslash\textbackslash}Reference [2]: Although \colorbox{yellow}{\textcolor{red}{Princeton}}{\textbf{\large{\color{green}\Checkmark}}} lays claim to the most college football national championships, Alabama leads the way since the introduction of the poll system in 1936. Whether the AP poll, BCS, or College Football Playoff, the Crimson Tide has won 13 national championships. With three titles, Nick Saban’s team was the most successful team of the BCS era. Furthermore, they’ve won the most college football titles under the current format.{\textbackslash\textbackslash}Reference [3]: - College Football Teams With the Most Championships Won{\textbackslash\textbackslash}Reference [4]: While Alabama is often considered the most successful college football program, the \colorbox{yellow}{\textcolor{red}{Princeton}}{\textbf{\large{\color{green}\Checkmark}}} Tigers have the most college football championships in the history of the sport. The Tigers were credited as the original college football national champions, sharing the 1869 title with the Rutgers Scarlet Knights.{\textbackslash\textbackslash}Reference [5]: Here are the eight teams who have won the most national championships in college football. \\
\hline
\textbf{Answer2:} & The \colorbox{yellow}{\textcolor{red}{Princeton}}{\textbf{\large{\color{green}\Checkmark}}} Tigers have the most college football championships in the history of the sport, with 10 titles[2][4]. Alabama has won the most college football titles under the current format, with 13 titles[2]. Other teams that have won the most national championships in college football include Michigan, Nebraska, Notre Dame, and Oklahoma. \\
\hline
\end{tabular}
\end{table*}

\section{Automatic Evaluation by GPT-4}
\label{sec:appendix7}
To enhance the assessment of the quality of our Question-Answer system, we conducted an Automatic Evaluation to evaluate the quality of our responses across multiple scoring dimensions. As shown in Table \ref{tab:Instructions of Automatic Evaluation for RAG by GPT-4}, GPT-4 was employed to compare and rank our method (CoV-RAG) against WebGLM in GLM-10b and Llama2-13b based on various scoring criteria, ranging from superior to inferior. The final ranking is shown in Table \ref{tab:accents}, and a case is shown in Table \ref{tab:Case of Automatic Evaluation by GPT-4}.

\begin{table*}[t]
\caption{Case of Winner Evaluation by GPT-4}
\center
\begin{tabular}{lp{\dimexpr\textwidth-25\tabcolsep-2\fboxrule}}
\hline
\label{tab:Case of Automatic Evaluation by GPT-4}\textbf{Question:} &  who is the first indian woman to be canonized as a saint \\
\hline
\textbf{Label:} & ["Saint Alphonsa"] \\
\hline
\textbf{Reference:} & Reference [1]: ...{\textbackslash\textbackslash}Reference [3]: \colorbox{yellow}{\textcolor{red}{Saint Alphonsa}}{\textbf{\large{\color{green}\Checkmark}}} was an Indian religious sister and educator. She was the first woman of Indian origin to be canonized as a saint by the Catholic Church. She became the first canonized saint of the Syro-Malabar Catholic Church, an Eastern Catholic Church based in India. Her feast day is observed on July 28th.{\textbackslash\textbackslash}... \\
\hline
\textbf{WebGLM10b:} & The first Indian woman to be canonized as a saint is Sister Alphonsa of Kerala[3]. She was an Indian religious sister and educator[3]. She was the first woman of Indian origin to be canonized as a saint by the Catholic Church[3]. Her feast day is observed on July 28th[3]. She was canonized by Pope Francis in 2015[3]. \\
\hline
\textbf{WebGLM13b:} & The first Indian woman to be canonized as a saint was Sister Alphonsa of Kerala[3]. She was an Indian religious sister and educator[3]. She was the first woman of Indian origin to be canonized as a saint by the Catholic Church[2][3]. Her feast day is observed on July 28th[3]. She was canonized by Pope Paul VI in October 1965[3]. \\
\hline
\textbf{CoV-RAG-S:} & The first Indian woman to be canonized as a saint was \colorbox{yellow}{\textcolor{red}{Saint Alphonsa}} of Kerala[3][4]. She was an Indian religious sister ... \\
\hline
\textbf{GPT-4 Eval:} &     rank Bias:  \colorbox{yellow}{\textcolor{red}{CoV-RAG-S}} > WebGLM13b > WebGLM-10b

    rank Citation Accuracy: \colorbox{yellow}{\textcolor{red}{CoV-RAG-S}} > WebGLM13b > WebGLM-10b
    
    rank Conciseness:  \colorbox{yellow}{\textcolor{red}{CoV-RAG-S}} >  WebGLM13b > WebGLM-10b

    rank Truthfulness:  \colorbox{yellow}{\textcolor{red}{CoV-RAG-S}}  > WebGLM13b = WebGLM-10b

    rank Correctness:   \colorbox{yellow}{\textcolor{red}{CoV-RAG-S}}  > WebGLM13b = WebGLM-10b \\
\hline
\textbf{Rank Reason:} & \textbf{Bias ranking reason}: CoV-RAG-S provides accurate information regarding Saint Alphonsa's canonization, including the correct Pope who canonized her, Pope Benedict XVI, in 2008. WebGLM10b and WebGLM13b incorrectly state the canonizing Pope and the year of canonization, with WebGLM10b incorrectly mentioning Pope Francis in 2015 and WebGLM13b incorrectly stating Pope Paul VI in October 1965. Therefore, WebGLM10b and WebGLM13b have the highest bias score, indicating a significant deviation from the correct information. 

    \textbf{Citation Accuracy ranking reason}: CoV-RAG-S correctly cites the information provided in the references without adding any incorrect details. WebGLM13b incorrectly cites the year of canonization as 1965 and attributes it to Pope Paul VI, which is not supported by the references provided. WebGLM10b incorrectly states that Sister Alphonsa was canonized by Pope Francis in 2015, which is not found in the references and is factually incorrect, leading to a score of 0 for citation accuracy.

    \textbf{Other scores ranking reason}: ...
\\
\hline
\end{tabular}
\end{table*}

\begin{table*}[t]
\caption{Instructions of Automatic Evaluation for RAG by GPT-4}
\label{tab:Instructions of Automatic Evaluation for RAG by GPT-4}
\center
\begin{tabular}{lp{\dimexpr\textwidth-25\tabcolsep-2\fboxrule}}
\hline
\textbf{Citation:} & Criteria Details For Answers include: \\
&\textbf{Citation Accuracy(0,1)}:  Evaluating whether the reference marks in the answer are accurate. \\
& \\

& \{
  \textbf{"question":} \textcolor{red}{(question)},
  \textbf{"reference":} \textcolor{red}{(reference)},
  \textbf{"answer1":} \textcolor{red}{(answer1)},
  \textbf{"answer2":} \textcolor{red}{(answer2)},
  \textbf{"answer3":} \textcolor{red}{(answer3)}
  \} \\

& \\
& Now you are a reading comprehension examiner who should do things as below: \\
& 1. Score the answer based on the evaluation criteria. \\
& 2. Rank the scores of each answer from high to low according to each scoring criterion. \\
& 3. Briefly state the reason for your Rank. \\
& \\
& \textbf{Output format example:} \\
& 
  {\{} 
"rank\_result": {\{}"Citation Accuracy": [("answer3", 0.77), ("answer1", 0.53), ("answer2", 0.12)]{\}}, "rank\_reason": "The reason for this ranking."
{\}} \\
  
\\
\hline
\textbf{Others:} & Criteria Details For Answers include: \\
&\textbf{Correctness(0,1)}:  Evaluating whether the question is correctly answered, you can refer to the golden label of the question below when evaluating. \\
&\textbf{Truthfulness(0,1)}:  Evaluating whether the text itself violates common sense, logic or contains contradictions. \\
&\textbf{Conciseness(0,1)}:  Evaluating whether the answer directly and succinctly addresses the question without unnecessary elaboration. \\
& \\

& \{
  \textbf{"question":} \textcolor{red}{(question)},
  \textbf{"golden label":} \textcolor{red}{(golden label)},
  \textbf{"answer1":} \textcolor{red}{(answer1)},
  \textbf{"answer2":} \textcolor{red}{(answer2)},
  \textbf{"answer3":} \textcolor{red}{(answer3)},
  \textbf{"answer4":} \textcolor{red}{(answer4)}
  \} \\

& \\
& Now you are a reading comprehension examiner who should do things as below: \\
& 1. Score the answer based on the provided evaluation criteria. \\
& 2. Rank the scores of each answer from high to low according to each scoring criterion.\\
& 3. Briefly state the reason for your Rank. \\
& \\
& \textbf{Output format example:} \\
& 
  {\{} 
"rank\_result": {\{}"Correctness": [("answer4", 0.77), ("answer1", 0.53), ("answer3", 0.37), ("answer2", 0.12)], "Truthfulness": [("answer3", 0.92), ("answer4", 0.41), ("answer2", 0.22), ("answer1", 0.02)], "Conciseness":[("answer4", 0.69), ("answer3",  0.51), ("answer1", 0.2), ("answer2", 0.15)]{\}}, "rank\_reason": "The reason for this ranking."
{\}} \\
\hline
\end{tabular}
\end{table*}

\begin{table*}[t]
\caption{Instruction of Automatic Evaluation for Revise by GPT-4}
\label{tab:Instructions of Automatic Evaluation for Revise by GPT-4}
\center
\begin{tabular}{lp{\dimexpr\textwidth-25\tabcolsep-2\fboxrule}}
\hline
\textbf{Instruction:} & Evaluate the appropriateness of revised questions and answers provided by four models. Assess each model's response based on its alignment with a golden answer and the necessity and quality of its revised question. \\
& 1. Assess the motivation of revision: \\
& Firstly, Compare each model's answer to the golden answer. Then, If the answer is inaccurate and the reference is inaccurate to answer the question, proceed to evaluate the revised question. Or, it's a poor revision timing. \\
& 2. Assess the content of revision. Note assess criterias are as followed: \\
&  (1). How well it improves content retrieval. \\
&  (2). Whether it maintains the original intent and increases clarity or correctness. \\
& \\
& Inputs: \\
& \{ \\
& \textbf{"Original Question"}: 
\textcolor{red}{(Original Question)}, 
  \textbf{"Golden Label":} \textcolor{red}{(Golden Label)},
  \textbf{"Reference":} \textcolor{red}{(Reference)},
  \textbf{"Model1":} \{"Answer1": \textcolor{red}{(Answer1)}, "Revised Question1": \textcolor{red}{(Revised Question1)}\},
  \textbf{"Model2":} \{"Answer2": \textcolor{red}{(Answer2)}, "Revised Question2": \textcolor{red}{(Revised Question2)}\} \\
&  \} \\

& \\
& Output Requirements: \\
& Rank the relvised questions based on their evaluation scores(threshold value of score should be between 0 and 1), from highest to lowest. Provide an overall reason for the ranking. \\
& \\
& Note you should only output the evaluate result, format is as followed: \\
& 
  {\{} 
"rank\_result": [{\{}"model": "1", "score": 0.9 {\}}, {\{}"model": "2", "score": 0.0 {\}}], 
"rank\_reason": "Overall Evaluation Reason"
{\}} \\
\hline
\end{tabular}
\end{table*}

\end{document}